\documentclass[10pt,twocolumn,letterpaper]{article}

\usepackage[pagenumbers]{cvpr} 

\usepackage{graphicx}
\usepackage{amsmath}
\usepackage{amssymb}
\usepackage{booktabs}
\usepackage{multirow}

\usepackage[pagebackref,breaklinks,colorlinks]{hyperref}

\usepackage[capitalize]{cleveref}
\crefname{section}{Sec.}{Secs.}
\Crefname{section}{Section}{Sections}
\Crefname{table}{Table}{Tables}
\crefname{table}{Tab.}{Tabs.}

\def\cvprPaperID{19} 
\def\confName{CVPR}
\def\confYear{2023}

\begin{document}

\title{TouchSDF: A DeepSDF Approach for 3D Shape Reconstruction \\using Vision-Based Tactile Sensing}

\author{
Mauro Comi$^{1}$ \and Yijiong Lin$^{1, 2}$ \and Alex Church$^{1,2}$ \and Alessio Tonioni$^{3}$ \and Laurence Aitchison$^{1}$ \and Nathan F. Lepora$^{1, 2}$\\ \\ $^{1}$University of Bristol \ \ \  $^{2}$Bristol Robotics Laboratory \ \ \  $^{3}$Google Zürich\\
{\tt\small mauro.comi@bristol.ac.uk}
}

\makeatletter
\let\@oldmaketitle\@maketitle 
\renewcommand{\@maketitle}{%
  \@oldmaketitle 
  \vspace{-1em}\centerline{\includegraphics[width=\textwidth]{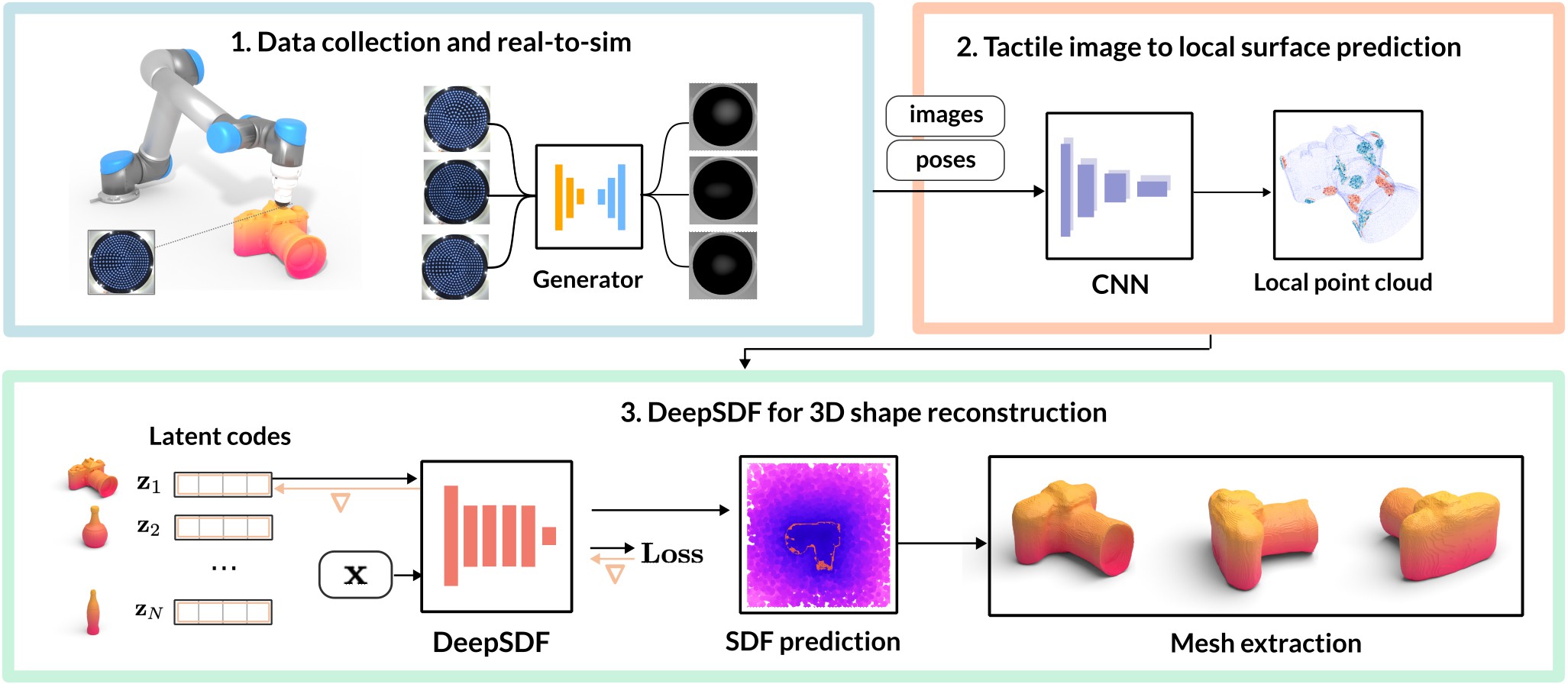}}
  \captionof{figure}{Overview of TouchSDF: (1) A robot samples the object's surface to obtain real tactile images (marker patterns) that are translated into simulated images (depth maps). (2) A Convolutional Neural Network (CNN) maps the simulated images to sets of 3D points representing the local object surface at the touch locations. (3) A pre-trained DeepSDF model predicts a continuous signed-distance function (SDF) representing the object shape from the point clouds over multiple contacts. The latent codes \(\mathbf{z_i}\) are 128-dimensional embeddings encoding the object's shape, while \(\mathbf{x}\) represents a 3D coordinate.}
  \label{fig:architecture_vertical}\bigskip
}
\makeatother

\maketitle

\begin{abstract}\vspace{-1em}
Humans rely on their visual and tactile senses to develop a comprehensive 3D understanding of their physical environment. Recently, there has been a growing interest in exploring and manipulating objects using data-driven approaches that utilise high-resolution vision-based tactile sensors. However, 3D shape reconstruction using tactile sensing has lagged behind visual shape reconstruction because of limitations in existing techniques, including the inability to generalise over unseen shapes, the absence of real-world testing, and limited expressive capacity imposed by discrete representations. To address these challenges, we propose TouchSDF, a Deep Learning approach for tactile 3D shape reconstruction that leverages the rich information provided by a vision-based tactile sensor and the expressivity of the implicit neural representation DeepSDF. Our technique consists of two components: (1) a Convolutional Neural Network that maps tactile images into local meshes representing the surface at the touch location, and (2) an implicit neural function that predicts a signed distance function to extract the desired 3D shape. This combination allows TouchSDF to reconstruct smooth and continuous 3D shapes from tactile inputs in simulation and real-world settings, opening up research avenues for robust 3D-aware representations and improved multimodal perception in robotics. Code and supplementary material are available at: \url{https://touchsdf.github.io/}
\end{abstract}

\section{Introduction}
\label{sec:intro}

The current state of 3D shape reconstruction research is primarily concerned with the sense of vision \cite{choy2016, tulsiani2017}. However, training Computer Vision algorithms for object manipulation is challenging due to the high-dimensional observation space, which suffers from occlusion and external lighting conditions. Methodologies based on high-resolution tactile sensing have recently been proposed as additional solutions for object exploration and manipulation \cite{yuan2017gelsight, lepora2021}. This type of perception presents several benefits over pure camera-based sensing, including the ability to capture contact-rich information, as well as being effective despite the presence of occlusions, homogeneous surfaces and transparent or reflective object materials. Additionally, tactile observations simplify the translation of simulated tasks to real-world scenarios as they require a smaller observation space \cite{church2021}.

Recently, several data-driven methodologies that utilise vision-based tactile sensors for 3D understanding have been proposed~\cite{higuera2022neural, rustler2022active, lepora2021}. To the best of our knowledge, Smith et al. \cite{smith2021} proposed the first and only approach for vision-based tactile reconstruction, which utilised DIGIT vision-based tactile sensors~\cite{lambeta2020} to extract contact-rich information. Although that work serves as a foundation for tactile-driven 3D reconstruction, a lack of testing in a real-world scenario makes it challenging to assess its applicability to robotic manipulation. Furthermore, the resulting reconstruction consists of separate meshes instead of smooth surfaces, potentially introducing discontinuities in the reconstructed shape. Notably, Smith et al. employ both touch-only and visuo-tactile data into their approach. For the purposes of our comparison, we focus on the touch-only model to investigate the possibility of 3D reconstruction using only tactile data, considering the advantages previously discussed.

In this work, we propose TouchSDF, a novel approach for 3D shape reconstruction that leverages an implicit neural function for vision-based tactile sensing. Unlike existing methods that utilise discrete representations, TouchSDF employs DeepSDF \cite{park2019deepsdf}, an implicit neural representation that encodes a smooth and continuous surface, enabling more accurate and robust reconstructions. DeepSDF has demonstrated the ability to reconstruct 3D geometry from partial point clouds, which is useful in our context as tactile sensors provide partial observations of an object. A key step for 3D reconstruction in the real world is to have effective real-to-sim tactile image translation, where we extend previous work with a GAN-based method \cite{church2021} to handle complex surfaces and 6D poses. Through evaluation on both simulated and real objects, we demonstrate the effectiveness of TouchSDF in tactile-based 3D shape reconstruction and highlight its potential in object exploration and manipulation tasks.

\textbf{Contribution} In summary, our work makes the following principal contributions:\\
\noindent 1) We propose TouchSDF, an approach for 3D shape reconstruction using vision-based tactile sensing. By leveraging the implicit neural representation DeepSDF, our method encodes smooth and continuous surfaces that enable accurate and robust reconstructions from partial observations.\\
\noindent 2) We demonstrate the ability to generalise across unseen objects and poses, both in simulation and in reality, by conditioning on latent variables, thus encoding a wide range of geometrical variations. \\
\noindent 3) We give the first evaluation of 3D shape reconstruction using purely vision-based tactile sensing on some \textit{real-world} 3D-printed and household objects.

\section{Related work}
Object reconstruction using tactile sensing is a new, active area of research that combines and adapts reconstruction methodologies from computer vision, graphics and robotics to a distinct perceptual modality: high-resolution touch.

The development of low-cost and open-source tactile sensors has opened new avenues of study in 3D reconstruction over the past few years \cite{meier2011,jia2010, yi2016, ilonen2014three}. To the best of our knowledge, \cite{wang20183d} proposed the first work on touch exploration for object reconstruction that employs vision-based tactile sensing (using a GelSight sensor \cite{yuan2017gelsight}). However, this technique yields a voxel-based reconstruction that is not always suitable for retrieving fine details. In contrast, Smith et al. \cite{smith2020, smith2021} proposed an approach that decouples vision and tactile sensing (using a DIGIT \cite{lambeta2020}) by leveraging a series of Graph Convolutional Networks to deform a default mesh into the desired object \cite{smith2019geometrics}. However, the lack of testing in the real world makes it challenging to assess the applicability of this reconstruction method in a robotic setting. In contrast, our approach results in a smooth and continuous surface reconstruction on both simulated and real-world shapes.

\begin{figure*}[t!]
\centerline{\includegraphics[width=1\textwidth]{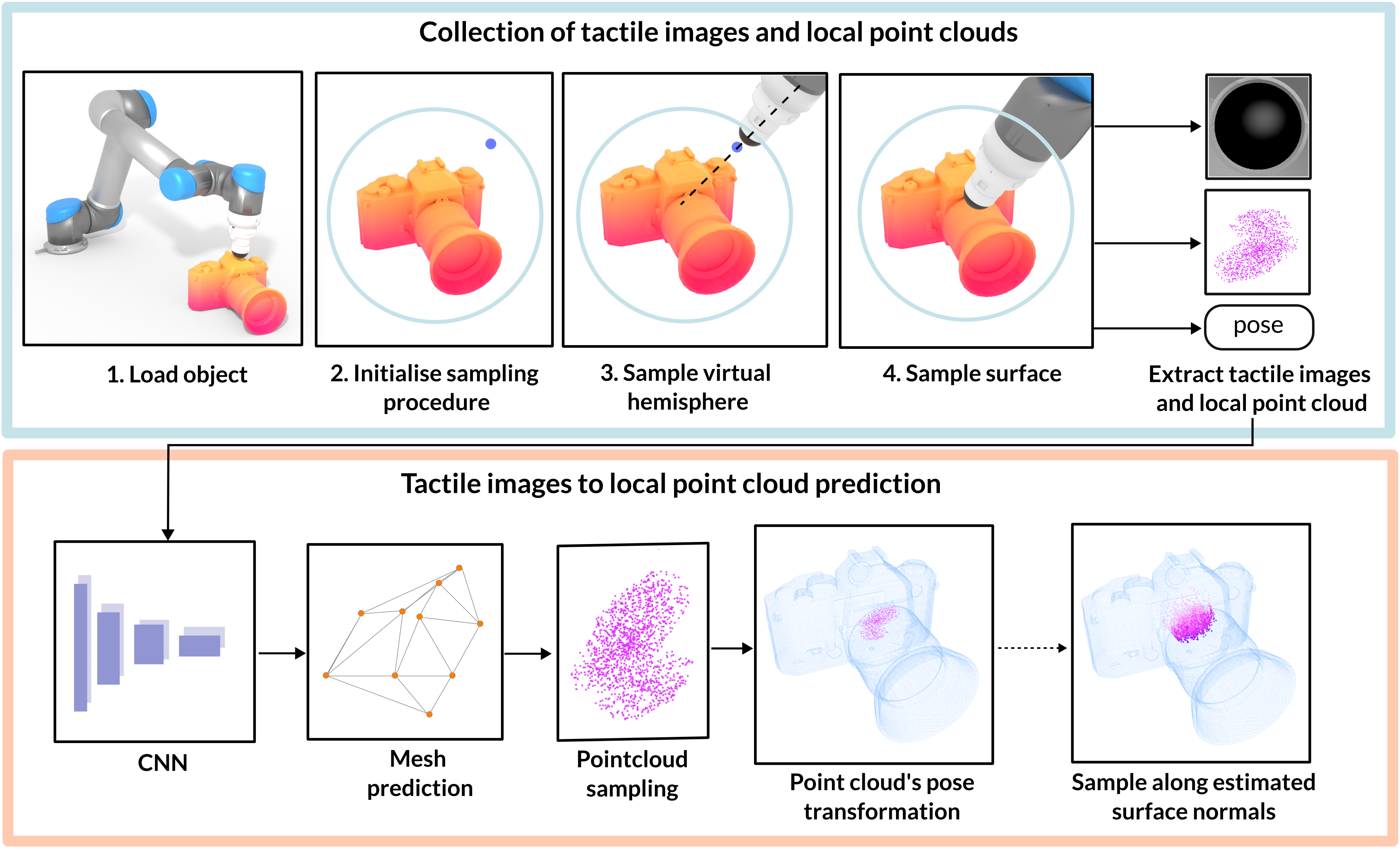}}
\caption{(Top) Data collection procedure for extracting the tactile image at the touch location, the sensor pose as an end effector to the robot arm, and local point cloud. (Bottom) Procedure to map a tactile image into a local point cloud with a CNN conditioned on the previously extracted data.}
  \label{fig:data_collection}
\vspace{-1em}
\end{figure*}

\textbf{Implicit neural representations in robotic manipulation.} Implicit Neural Representations (INRs), such as DeepSDF \cite{park2019deepsdf}, Occupancy Networks \cite{mescheder2019occupancy} and NeRF \cite{mildenhall2021nerf}, have recently been proposed as state-of-the-art approaches for many tasks in graphics, vision and robotics. In particular, numerous works have explored the use of INRs in manipulation tasks \cite{higuera2022neural, driess2022learning, wi2022virdo, simeonov2022neural, fu2023neuse, jiang2021synergies}.  However, while NeRF-based approaches excel in view-synthesis tasks, they are not suited for reconstruction tasks based purely on tactile sensing, as the output produced by tactile sensors is a local depth map rather than a global view of the scene. Consequently, our proposed approach relies on DeepSDF, an efficient INR for encoding and reconstructing 3D surfaces using solely geometric information.

\textbf{Tactile sim-to-real.} Physics engines approximate sensor deformation, resulting in a disparity between real and simulated tactile images. Numerous approaches have been suggested to bridge this simulation-to-reality gap, such as finite-element methods \cite{narang2021sim, sferrazza2020sim} and rigid-body physics approximation \cite{church2021, agarwal2021simulation, wang2022tacto}. Recently, Higuera et al. \cite{higuera2023learning} proposed a diffusion model-based technique \cite{sohl2015deep} to generate realistic tactile images from simulated contact depths. However, diffusion models have high computational requirements which limit their suitability in real-time and low resource environments. Hence, we follow the approach proposed by \cite{church2021} in utilising a fast rigid-body physics simulator to simulate depth images and a one-step image translation model trained using a Generative Adversarial Network (GAN) \cite{goodfellow2014generative}, which is more suitable for real-time robotics applications.

\section{Methodology}
\label{sec:methodology}
Our reconstruction procedure has three steps (\cref{fig:architecture_vertical}). In the first step, we collect tactile images by sampling the surface of the object we intend to reconstruct. We also store the sensor (end-effector) pose at contact alongside the local point cloud describing the contact geometry. Real-world tactile images are converted into simulated tactile images using a real-to-sim GAN model \cite{church2021}. In the second step, we predict the contact geometry based on the simulated tactile image and sensor pose. Finally, we employ a pre-trained DeepSDF model to reconstruct objects using the predicted contact geometry (local point cloud). Both our \textit{contact geometry prediction} and \textit{DeepSDF} models were trained separately on objects sampled from the ShapeNet and ABC datasets. While ShapeNet consists of 3D models of various object categories, the ABC dataset lacks categorical divisions and includes objects synthetically generated to encompass a wide range of geometries.

\subsection{Simulation}\label{sec:simulation}
\textbf{Collection of tactile images and local point cloud.} Tactile images, local point clouds and sensor poses were collected for training purposes using a PyBullet-based tactile simulator (Tactile Gym \cite{church2021} \cite{lin2022}). The data collection procedure involved four steps. First, the objects to be reconstructed are loaded into our simulator and the sampling domain is automatically restricted to the shape boundaries of the object. Second, a virtual sphere is created around the object (\textit{Sample surface} in Fig. \ref{fig:data_collection}), from which a random location is sampled. Third, the robot arm moves the end effector-mounted tactile sensor to this location while keeping the sensor oriented towards the centre of the object. Last, as the sensor touches the object, the surface contact area is transduced into a 256$\times$256 pixel tactile image. This image is obtained by rendering the contact depth at the touch location using the PyBullet's synthetic camera. To control the touch depth, the average image intensity $\mathrm{D}_t \in [0, 255]$ is checked every timestep $t$; then the sampling procedure is stopped when $\mathrm{D}_t > 1$. In addition to the tactile image, the sensor's pose and a local point cloud are extracted. This point cloud serves as the ground truth to train a model that maps a tactile image into the corresponding touched surface.

\textbf{Tactile images to point cloud prediction.} We adapted the approach of Smith et al. \cite{smith2021} to learn a mapping between tactile images and local point clouds. Our data collection involved a simulated 6-DoF robot arm equipped with a simulated TacTip tactile sensor~\cite{church2021,lin2022} to sample the surfaces of the objects used for training and validation. To train our local chart prediction model, we gathered single-channel 2D tactile images in simulation, with each image representing the depth map of the touched area on the object's surface $\mathcal{S}$. Additionally, we extracted a ground truth point cloud of the surface at the touch location. Our objective is to employ a Convolutional Neural Network (CNN) to predict the contact geometry by deforming a base mesh. To achieve this, we followed the procedure outlined in \cite{smith2021} and shown in Fig.~\ref{fig:data_collection}. Initially, we define a base triangular mesh with 25 fixed vertices and faces, serving as the initial state for subsequent deformations. Then, a CNN receives a tactile image as input and produces the relative displacement for each vertex in the base mesh, which is deformed by displacing each vertex according to its corresponding predicted displacement value. Next, we uniformly sample a point cloud on the deformed mesh, generating a set of points representing the shape at touch location. Additionally, we enhanced the point cloud by estimating the normals of the surface and sampling $m$ additional points along them, both in the positive and negative directions. Finally, the CNN is optimised to minimise the Chamfer Distance \cite{sun2018pix3d} between the point cloud sampled on the predicted mesh and the ground truth point cloud.

\textbf{3D shape reconstruction.}
DeepSDF \cite{park2019deepsdf} is an implicit neural representation that encodes 3D shapes as Signed Distance Functions (SDF), enabling smooth shape reconstruction from partial observations. In this work, we utilised DeepSDF to reconstruct 3D shapes, leveraging partial object surfaces estimated from tactile sensor observations. Following the methodology outlined in \cite{park2019deepsdf} on DeepSDFs, we constructed a dataset of SDF pairs $\{(\mathbf{x_j}, s_j)\}_{j=1}^N$ per object, where $\mathbf{x} \in \mathbb{R}^3$ are the coordinates sampled within a closed volume around the objects and $s_j \in \mathbb{R}$ corresponds to the associated signed distances. We adopted the architecture introduced by \cite{park2019deepsdf} with added positional encoding $\gamma(\cdot)$ \cite{mildenhall2021nerf, vaswani2017attention}, where \mbox{$\gamma: \mathbb{R}^3 \rightarrow \mathbb{R}^m$} is a Fourier feature representation encoding each spatial coordinate $\mathbf{x_j}$ into a set of high-frequency sinusoidal functions $\gamma(\mathbf{x_j}) = [\gamma_1(\mathbf{x_j}), \gamma_2(\mathbf{x_j}), ..., \gamma_m(\mathbf{x_j}) ]$. Positional encoding, which we use in place of raw 3D coordinates, has proved to be extremely efficient in encoding high-frequency functions \cite{tancik2020fourier}, as demonstrated in recent advancements in neural radiance fields \cite{mildenhall2021nerf}. This aligns with recent findings \cite{rahaman2018spectral} indicating that deep networks tend to bias towards lower frequency functions. Positional encoding projects inputs into a higher-dimensional space using high-frequency functions, thus improving the model's ability to capture high-frequency variations in the data. During training, the DeepSDF model learns a continuous signed distance function $f_{\theta}(\mathbf{x}, \mathbf{z})$ conditioned on a shape embedding $\mathbf{z}$. To ensure the encoding of shape information, an embedding $\mathbf{z_i}$ is jointly optimized with the network parameters $\theta$ for each shape $i$. As a result, similar shapes are encoded by similar latent vectors, enabling interpolations in the latent space that facilitate the reconstruction of unseen objects. The objective minimised during the training process follows \cite{park2019deepsdf}:
\begin{equation}\label{eq:loss}
    \mathcal{L}(\theta, \mathbf{z}) = \underset{\theta, \mathbf{z_i}}{\text{arg min}} \sum_i \sum_{j=1}^N || f_{\theta}(\gamma\mathbf{(x_j)}, \mathbf{z_i}) - s_j||_1 + \alpha \cdot ||\mathbf{z_i}||_2^2
\end{equation}
Here the regularisation term $||\mathbf{z_i}||_2^2$ weighted by $\alpha$ is crucial to avoid exploding gradients when optimising latent vectors. Additionally, this term serves as a prior on the shape embeddings, enhancing their retrieval during inference.

\textit{Reconstruction.} The CNN-predicted point cloud $\mathcal{O}$ provides a partial observation of the geometry at the touch location, including samples along the normals of the estimated contact surface. Points directly on the surface are assigned a signed distance value of zero, while those along the normals receive positive or negative distances. This approach enhances the robustness of the DeepSDF model at inference time, as it provides not only surface points but also samples inside and outside the shape.  To reconstruct the complete shape of the target object, a DeepSDF auto-decoder \cite{park2019deepsdf} is conditioned on $\mathcal{O}$ to optimise a 128-dimensional latent code $\mathbf{z_i}$
by solving $\text{arg max}_{\mathbf{z_i}} \mathbb{P}(\mathbf{z_i} | \mathcal{O})$ via first-order optimisation. This is achieved by freezing the model's parameters $\theta$ and solving $\text{arg min}_{\mathbf{z_i}} \mathcal{L}(\mathbf{z_i})$. The inferred \textit{global} latent vector represents a compact representation of the entire shape that best describes the partial observation. Inspired by GAN inversion \cite{roich2022pivotal}, we additionally freeze the latent vector and fine-tune the model's parameters on the partial observation. The signed-distance function is defined by conditioning a pre-trained DeepSDF model on coordinates $\mathbf{x}$ and inferred latent vector $\mathbf{z_i}$. Finally, a surface $\mathcal{S}$ is extracted by employing the deterministic Marching Cubes algorithm \cite{lorensen1987marching}, which extracts the zero-level set of the predicted signed-distance function \mbox{$\mathcal{S} = \{\mathbf{x} \in \mathbb{R}^3 | f_{\theta}(\mathbf{x}, \mathbf{z}) = 0 \}$}.

\subsection{Real world}\label{subsect:real_world}

\textbf{Hardware.} We conducted real-world experiments with a 6-DoF industrial robot arm (ABB IRB 120). The robot was equipped with a high-resolution vision-based tactile sensor, for which we used a TacTip 3D-printed soft biomimetic tactile sensor \cite{lepora2021}. For real-world evaluation, we 3D-printed four objects (two different bottles, a camera, and a bowl) selected from ShapeNetCore.V2 \cite{shapenet2015} and selected two household objects (a transparent jar and a mug). Note that all these objects are unseen during training. The 3D-printed objects were designed with mounting sockets bolted onto a table as a secure fixture during tactile interactions and the everyday objects were fixed with a clamp.

\textbf{Real-to-Sim Image Transfer.} The point cloud prediction model, which predicts the coordinates that DeepSDF uses for shape completion, is trained on simulated tactile images. These exhibit significant deviations from real tactile images (Fig. \ref{fig:architecture_vertical}, top-left box). When reconstructing objects in the real world, the robot collects real tactile images, which the point cloud prediction model is not trained to process. Therefore, we map the real tactile images to simulated ones using the translation approach proposed in \cite{church2021}, which utilizes a Generative Adversarial Network (GAN) framework for image-to-image translation. The pix2pix GAN~\cite{isola2017image} is trained with pairwise simulated (depth map) and real tactile images (marker patterns) collected from the objects set introduced in \cite{gomes2021generation} and not part of the test data. We adopted the hyperparameters from \cite{church2021}, which gave good performance in the translation process \cite{church2021} \cite{lin2022}.

\textbf{Sim-to-Real Object Reconstruction.} In this step, we combine the point cloud prediction model, DeepSDF and real-to-sim image transfer to achieve sim-to-real object reconstruction. We collect real tactile images by performing random contacts with the robotic arm-mounted tactile sensor onto the real-world objects, using the same contact criteria as the simulated data gathering (Sec. ~\ref{sec:simulation}). For each touch, the real image is translated into a simulated tactile image and labeled with the corresponding end effector pose. The CNN is used to predict a local point cloud describing the contact geometry. Our method is evaluated by computing the mean absolute error between the predicted point clouds and the object (Sec. \ref{subsect:sim-to-real}), which is possible due to the availability of CAD models for those 3D-printed objects. Finally, our pre-trained DeepSDF model is conditioned on the predicted point clouds and estimated signed distance to extract the shape of an object (see Sec. \ref{sec:simulation}).


\begin{figure*}[h!]
  \centering
  \includegraphics[width=0.9\textwidth]{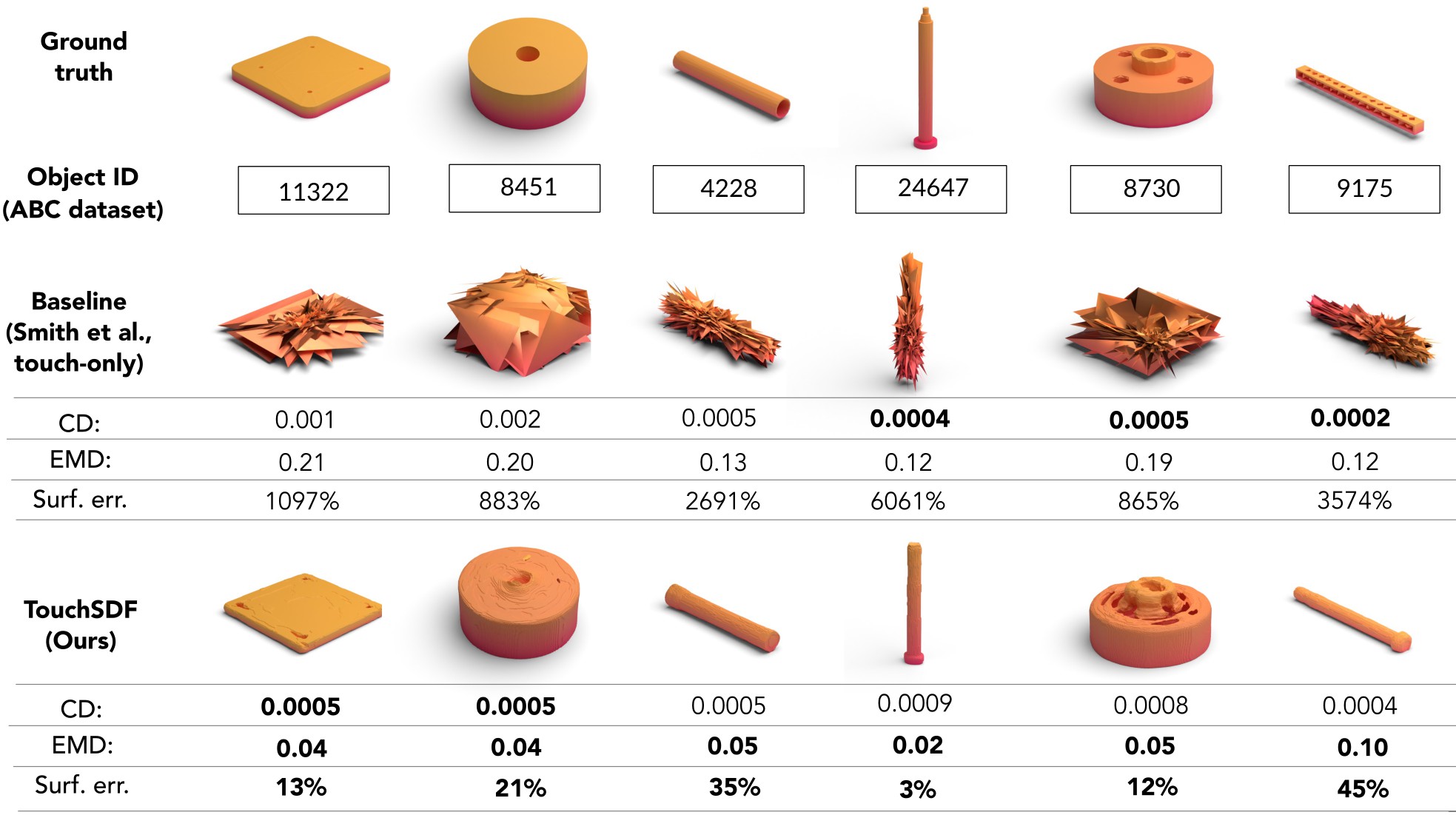}
  \caption{Comparison of TouchSDF and Smith et al. (grasp-only) performance after 20 touches. TouchSDF achieves a better score on EMD, Surface error, and visual quality, despite occasionally registering lower CD values.}
  \label{fig:ABC_visual_results}
\end{figure*} 

\section{Results}
\label{sec:result}

\subsection{Metrics}

Our experimental evaluation utilised three metrics: Chamfer Distance \cite{sun2018pix3d}, Earth Mover's Distance (EMD) \cite{fan2017point}, and Surface error.
While CD is a common metric in 3D reconstruction tasks, it has documented limitations in accurately measuring the visual quality of reconstructed meshes \cite{wang2018pixel2mesh} \cite{tatarchenko2019single}. This limitation stems from CD's focus on point-wise distance rather than on the distribution of the predicted point cloud. This can be observed in Fig. \ref{fig:ABC_visual_results}, where shapes with lower visual quality reconstructions achieve better (lower) CD scores. To address this shortcoming, EMD optimises a mapping that minimises the average distance between corresponding points in the predicted and ground truth point sets. Because EMD focuses on the entire point cloud distribution, it is more faithful to visual quality. However, these metrics do not provide a measure of the average error across surface reconstruction, which is informative in assessing the quality of a reconstructed mesh. Therefore, we introduced the Surface Error, which quantifies the average error on the surface reconstruction relative to each mesh. Specifically: 
$$ \text{Surface error (\%)} = \frac{|S_{p} - S_{gt}|}{S_{gt}} \cdot 100 $$
where $S_{gt}$ and $S_{p}$ are the surface of the ground truth and predicted mesh respectively.

\subsection{Simulation}
\begin{figure*}
  \centering
  \includegraphics[width=1\textwidth]{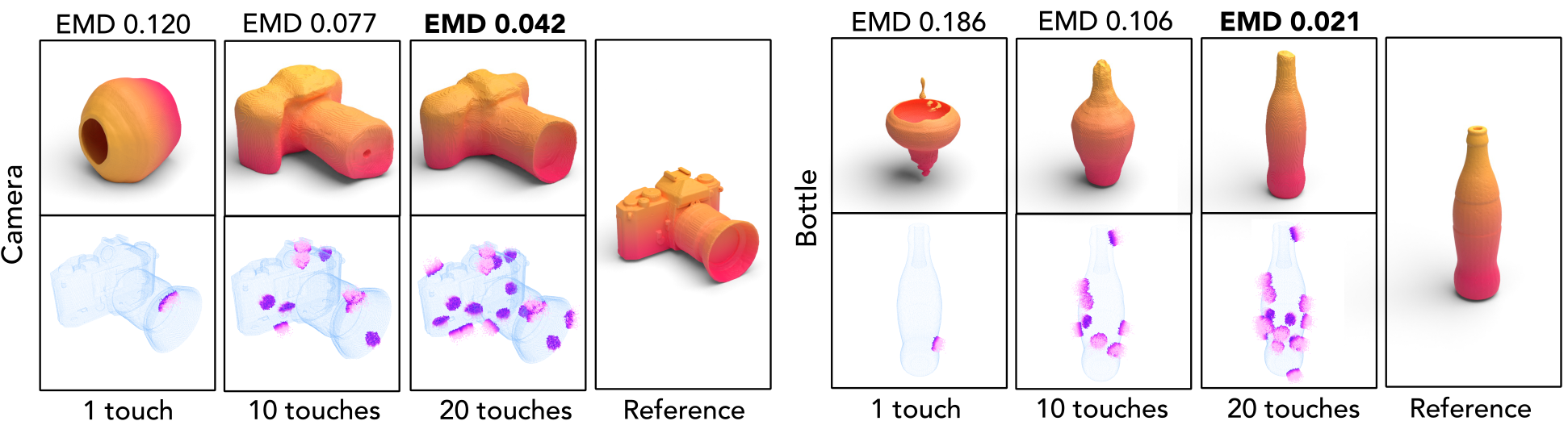}
  \vspace{-1em}
  \caption{Reconstruction results of TouchSDF for 1, 10 and 20 touches. The Chamfer Distance (CD) between the reconstructed and ground truth meshes is reported. Predicted point clouds referring to contact geometries are shown in purple (darker tones represent positive and lighter tones negative distances).}
  \label{fig:reconstruction_matrix}
\end{figure*}

We trained and tested our models using objects from the ABC and the ShapeNet \cite{shapenet2015} datasets. While for the ABC dataset we were able to benchmark our model's performance against the \textit{grasp-only} model proposed by Smith et al., a direct comparisons on the ShapeNet dataset was not performed as Smith et al.'s evaluation and provided model were not tailored to this specific dataset. However, our results could serve as a benchmark for future assessments.

\textbf{ABC dataset.} We sampled 3500 shapes for training, 350 for validation and 200 for testing. To ensure a fair comparison with Smith et al. \cite{smith2021}, shapes were randomly sampled from the training, validation, and testing sets used by those authors. Table \ref{tab:ABC_results} shows the CD, EMD and Surface error obtained by TouchSDF and Smith et al. after 20 touches. To evaluate their approach, we used the \textit{grasp-only} model provided by the authors. Both EMD and Chamfer Distance were computed using 4096 points. TouchSDF achieves better EMD and surface reconstruction error, but slightly lower CD despite better visual quality (Fig.~\ref{fig:ABC_visual_results}). It is important to highlight that, in contrast to our approach, Smith et al.'s reconstruction model was trained to explicitly minimise the CD between predicted and ground truth meshes.

\begin{table}[h]
    \small
    \centering
    \begin{tabular}{c c c c }
    \toprule
     & CD ($\downarrow$) & EMD ($\downarrow$) & Surface error ($\downarrow$) \\ 
     \midrule   
     Smith et al. & \multirow{2}{*}{\textbf{0.003}} & \multirow{2}{*}{0.19} & \multirow{2}{*}{2785\%} \\
     (grasp-only) &  &  & \\
     TouchSDF (ours) & 0.006 & \textbf{0.07} & \textbf{36\%}  \\
    \bottomrule
    \end{tabular}
    \caption{Comparison of TouchSDF and Smith et al. after 20 touches. }
    \label{tab:ABC_results}
    \vspace{-2em}
\end{table}

\begin{figure*}[t!]
  \centering
  \includegraphics[width=0.9\textwidth]{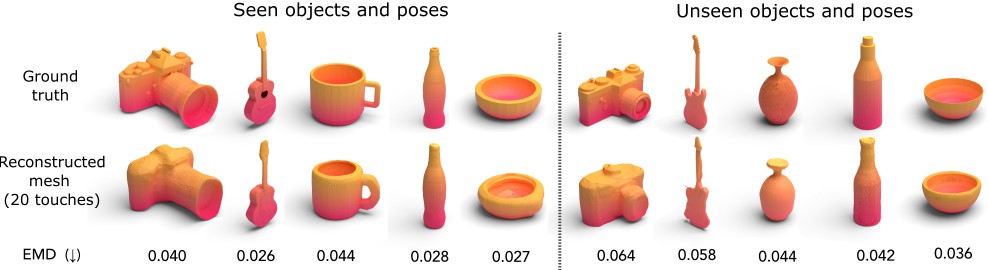}
  \caption{Reconstructions of seen and unseen objects (ShapeNet). The consistent EMD scores demonstrate TouchSDF's ability to generalise across categories.}
  \label{fig:reconstruction_touch}
\end{figure*} 

\textbf{ShapeNet dataset.} We collected a dataset consisting of 1,650 objects, which were then divided into three subsets: 1,100 objects for training, 200 for validation and 350 for testing. The chosen objects span six diverse categories considered previously in related work: bowls, bottles, cameras, jars, guitars and mugs. \cref{fig:reconstruction_matrix} and \cref{tab:seen_unseen} show an improvement in reconstruction quality as the number of touches increases. This results from a higher coverage of the object surface. Fig. \ref{fig:reconstruction_matrix} displays a set of point clouds predicted from tactile images. We present the object point clouds of the reference shape (in blue) solely for visualisation purpose, as our model exclusively relies on the predicted point clouds (in purple). By leveraging points with both positive and negative distance, we increased the robustness of the latent code inference procedure.

\begin{table}[h]
    \small
    \centering
    \begin{tabular}{c c c c}
    \toprule
    & \multicolumn{3}{c}{Unseen objects and poses} \\ 
    \# Touches & 1 & 10 & 20  \\
    \midrule
    
    Bottle & 0.113 \tiny{$\pm$ 0.085} & 0.082 \tiny{$\pm$ 0.058} & \textbf{0.047 \tiny{$\pm$ 0.024}}  \\
    
    Mug & 0.091 \tiny{$\pm$ 0.030} & 0.072 \tiny{$\pm$ 0.024} & \textbf{0.066 \tiny{$\pm$ 0.018}}  \\
    
    Bowl & 0.085 \tiny{$\pm$ 0.034} & 0.073 \tiny{$\pm$ 0.029} & \textbf{0.048 \tiny{$\pm$ 0.017}}\\
    
    Camera & 0.131 \tiny{$\pm$ 0.035} & 0.101 \tiny{$\pm$ 0.046} & \textbf{0.092 \tiny{$\pm$ 0.043}} \\
    
    Guitar & 0.195 \tiny{$\pm$ 0.102}  & 0.177 \tiny{$\pm$ 0.104} & \textbf{0.155 \tiny{$\pm$ 0.087}} \\
    
    Jar & 0.164 \tiny{$\pm$ 0.113} & 0.136 \tiny{$\pm$ 0.107} & \textbf{0.071 \tiny{$\pm$ 0.038}} \\
    \midrule
    Average & 0.136 \tiny{$\pm$ 0.092} & 0.112 \tiny{$\pm$ 0.086} & \textbf{0.081 \tiny{$\pm$ 0.062}} \\

    \bottomrule
    \end{tabular}
    \caption{EMD for unseen objects and poses in simulation (lower is better). Results are reported for 1, 10 and 20 touches. Reconstruction quality improves with number of touches. }
    \label{tab:seen_unseen}
    \vspace{-1em}
\end{table}

To assess the accuracy of the reconstructed shapes, we considered 300 unseen objects and poses across the six distinct categories previously mentioned, using 1, 10, and 20 touches per object. To measure the reconstruction quality, we computed the EMD between predicted and ground truth objects and then computed the average metrics for each category (Table \ref{tab:seen_unseen}). Our results indicate that increasing the number of touches leads to much better reconstruction quality across all categories. The \textit{Guitar} category has a higher EMD than average, likely due to the substantial geometric variations in objects belonging to this category. As a result, it is challenging for the auto-decoder framework to optimise robust global shape embeddings for these objects.

Moreover, the relatively low difference in EMD scores for seen and unseen objects obtained for 20 randomly-sampled touches (Fig. \ref{fig:reconstruction_touch}) shows that our model is able to generalise effectively to unseen objects and poses within known categories. The obtained reconstructions retain details that could be useful for downstream tasks in robot learning, such as cup handles and concave surfaces.

\subsection{Real world}\label{subsect:sim-to-real}


\begin{figure*}[h!]
  \centering
  \includegraphics[width=.8\textwidth]{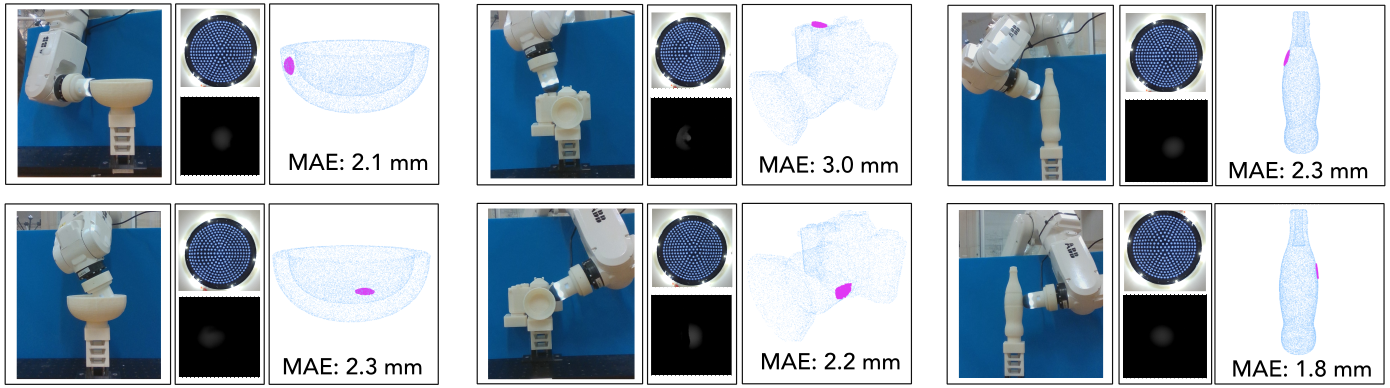}
  \caption{Examples of real-world touches, with the corresponding real and simulated tactile images, and predicted point cloud superimposed on the object, for the 3D-printed camera, bowl and bottle.} 
  \label{fig:real2sim}
\end{figure*} 

\begin{figure*}[h]
  \centering
  \includegraphics[width=.8\textwidth]{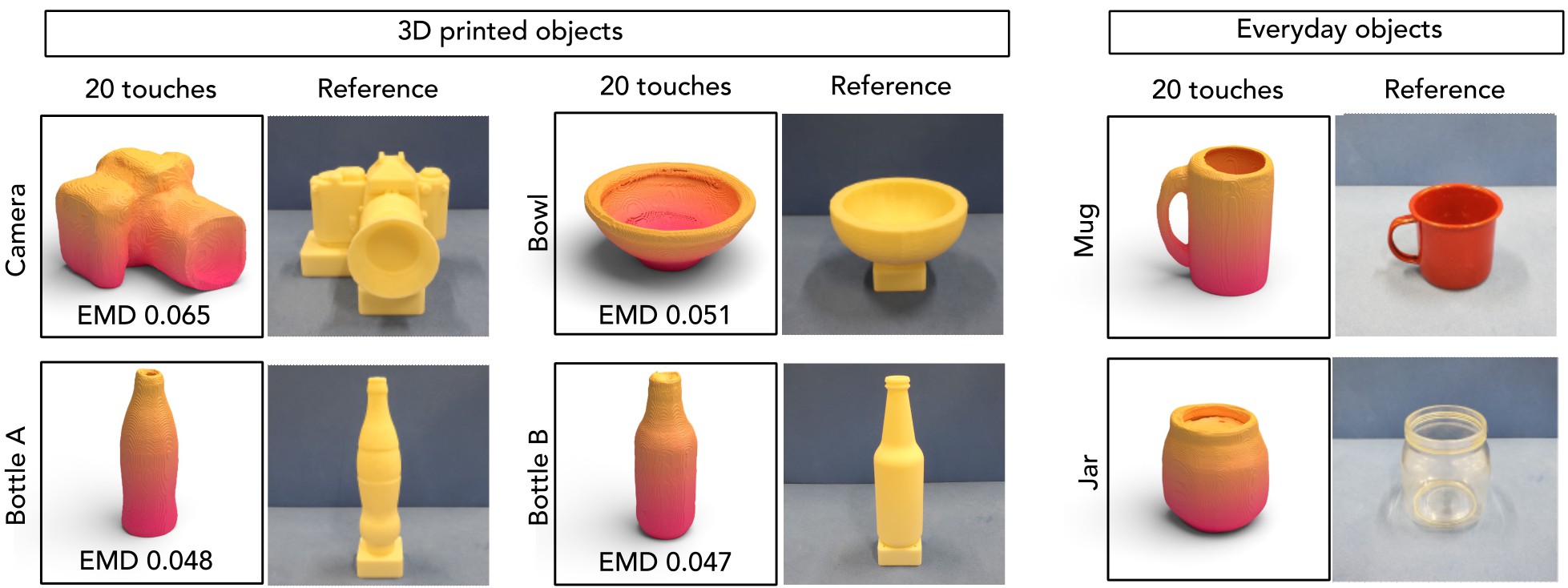}
  \caption{Reconstruction of four real 3D-printed objects and two everyday objects. The Earth Mover's Distance (EMD) calculation requires a CAD model, hence it is applicable to 3D-printed objects but not to everyday objects. The cube-shaped mounting sockets are not part of the objects.}
  \label{fig:reconstruction_real}
\end{figure*}

\textbf{Real-to-Sim.} To evaluate our sim-to-real procedure, we collected 160 real-world tactile images from the four 3D-printed objects illustrated in Fig. \ref{fig:reconstruction_real}. We employed the Real-to-Sim Image Transfer approach described in Section \ref{subsect:real_world} to convert these real tactile images into simulated tactile images. Next, our local surface prediction CNN was utilised to predict local point clouds based on the translated tactile images. The evaluation of our method involved calculating the Mean Absolute Error (MAE) between the predicted local point clouds and the corresponding contact geometries obtained from our tactile simulator at the touch location. The results (Table \ref{tab:sim2real}) demonstrate that our method achieved an average error from 2-4\,mm over the considered objects when predicting contact geometry. Figure \ref{fig:real2sim} shows examples of collected tactile images, their translations from real to simulated images, and the resulting predicted local point clouds superimposed onto the original CAD model. 

\begin{table}[!h]
\small
\centering
\begin{tabular}{c c c c c }
\toprule
 & Camera & Bowl & Bottle A & Bottle B \\ 
 \midrule
 MAE (mm) & 4.0 & 3.1 & 2.5 & 2.3 \\
\bottomrule
\end{tabular}
\caption{Mean average error between predicted local point clouds and contact geometries.}
\label{tab:sim2real}
\end{table}

\textbf{Reconstruction.} To reconstruct the 3D-printed objects, we collected real tactile images from random locations on the object surfaces and translated them into their corresponding simulated images. The reconstruction procedure followed the same methodology employed in simulation. As the embeddings optimised by DeepSDF are not $SE(3)$-invariant, it was necessary to ensure that the pose of the object being reconstructed is consistent with the poses observed during training. To achieve this, the world reference frame was set manually to the centre of the objects. Our method successfully reconstructed both real 3D-printed objects and additional everyday objects (a mug and a transparent jar), achieving low EMD values that are comparable to those obtained in simulation (Fig. \ref{fig:reconstruction_real}). The EMD could not be computer for everyday objects to the lack of CAD models. 

\section{Discussion and limitations}
\label{sec:discussion}
In this study, we introduced TouchSDF, a new approach for reconstructing 3D objects using tactile images captured by a high-resolution vision-based tactile sensor (the TacTip), both in simulation and in reality. To the best of our knowledge, this is the first work capable of reconstructing smooth and continuous 3D objects in the real world using solely tactile sensing. Our work was enabled by advancements in real-to-sim translation techniques for vision-based tactile images~\cite{church2021} coupled with progress in neural implicit representations~\cite{park2019deepsdf}. By leveraging TouchSDF, which was trained entirely in simulation, we successfully reconstructed real-world objects via real-to-sim translation. In combination with a CNN capable of mapping tactile images to local point clouds, we obtained precise local surface reconstructions with an average error of around 2-4\,mm. Additionally, we demonstrated that the reconstruction quality improves notably with increasing the number of touches. Overall, we expect that our approach can be adapted to various types of vision-based tactile sensors, as the real-to-sim technique we used here has also been shown to work well with a GelSight-type sensor \cite{lin2022}.

\textbf{Limitations and future directions}. Because DeepSDF is not inherently $SE(3)$-invariant, the performance of our method relies on the object's pose and the ability to interpolate in the latent space to generalise over unseen poses. Moreover, objects with unique shapes and features are harder to reconstruct, as retrieving their corresponding latent vector at test-time optimisation is challenging. Potential solutions involve augmenting the dataset with objects in different poses and exploring $SE(3)$-invariant methodologies for shape completion using DeepSDF. Another limitation lies in the use of an auto-decoder \cite{park2019deepsdf} within the DeepSDF framework, which results in slow latent code optimisation. The implementation of a standard encoder-decoder architecture could potentially expedite the process, as it does not rely on the optimisation of a latent code at test time. Furthermore, the random selection of touch locations is motivated by the lack of initial knowledge regarding the object to be reconstructed, and it would be interesting to explore strategies for inferring the next optimal touch location that maximises reconstruction quality as in~\cite{smith2021}. Lastly, a potential avenue for future research in 3D understanding lies in integrating vision with tactile sensing. By leveraging the rich information captured by both modalities, we would expect that the techniques could be extended to enable sophisticated object exploration and manipulation tasks.

{\small
\bibliographystyle{ieee_fullname}
\bibliography{egbib}
}

\end{document}